%% file: dacy.tex
\documentclass{article}

\usepackage{arxiv}

\usepackage[T1]{fontenc}    
\usepackage[usenames,svgnames,dvipsnames]{xcolor}
\usepackage[colorlinks,linktoc=all]{hyperref}
\usepackage[all]{hypcap}
\hypersetup{citecolor=DarkSlateGrey}
\hypersetup{linkcolor=DarkSlateGrey}
\hypersetup{urlcolor=DarkSlateGrey}
\usepackage{url}            
\usepackage{booktabs}       
\usepackage{amsfonts}       
\usepackage{nicefrac}       
\usepackage{microtype}      
\usepackage{lipsum}		
\usepackage{graphicx}
\usepackage[style=apa]{biblatex}
\usepackage{doi}
\usepackage{longtable}
\usepackage{array}
\usepackage{multirow}
\usepackage{wrapfig}
\usepackage{float}
\usepackage{colortbl}
\usepackage{pdflscape}
\usepackage{tabu}
\usepackage{threeparttable}
\usepackage{threeparttablex}
\usepackage[normalem]{ulem}
\usepackage{makecell}

\usepackage[nopostdot]{glossaries} 
\usepackage{glossary-mcols} 
\makeglossaries
\newacronym{ner}{NER}{Named Entity Recognition}
\newacronym{nlp}{NLP}{Natural Language Processing}
\newacronym{pos}{POS}{Part-of-Speech}
\usepackage[nameinlink]{cleveref}

\title{\includegraphics[scale=0.06]{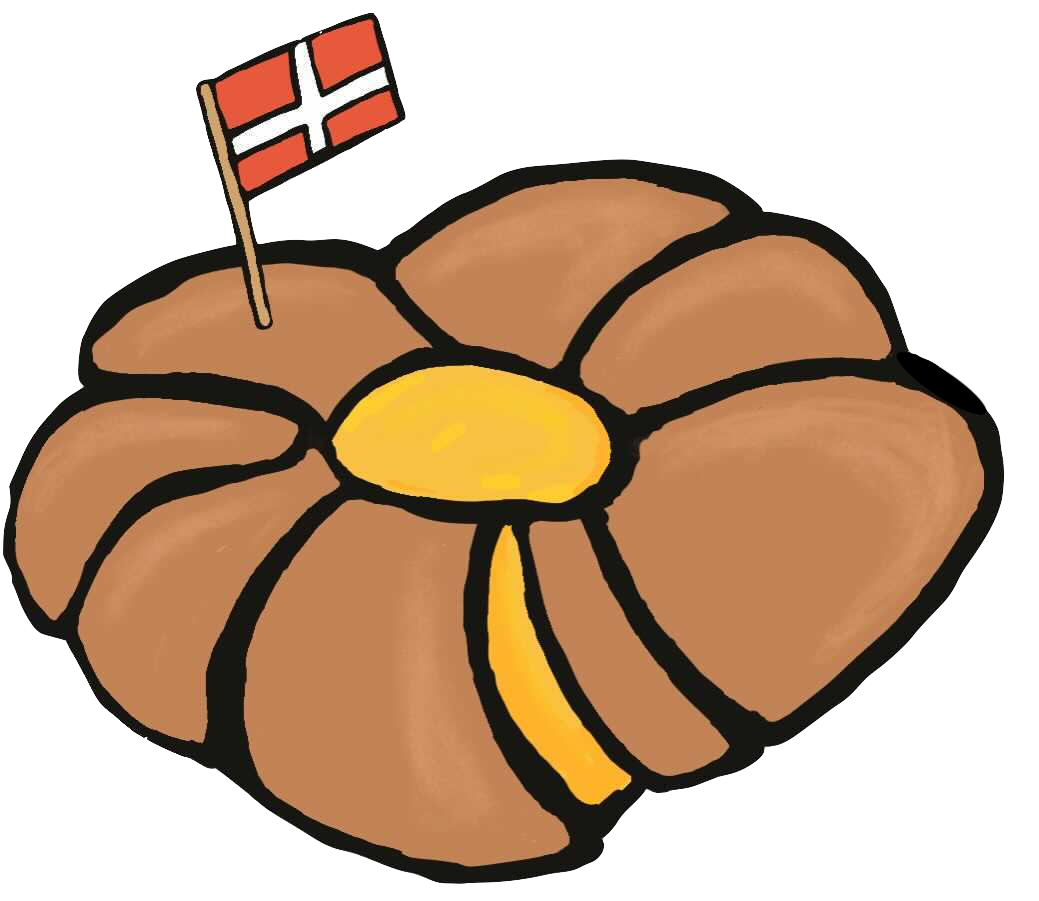}
DaCy: A Unified Framework for Danish NLP}

\author{ \href{https://orcid.org/0000-0001-8733-0966}{\includegraphics[scale=0.06]{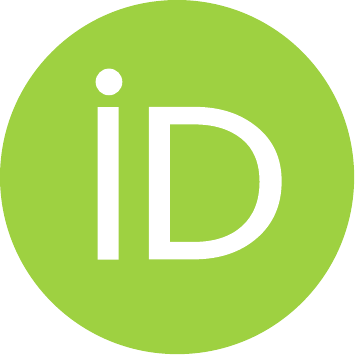}\hspace{1mm}Kenneth C.~Enevoldsen}\\
	Interacting Minds Centre \&\\
	Center for Humanities Computing Aarhus\\
	Aarhus University\\
	Jens Chr. Skous Vej 4, Building 1483, 3rd floor\\
    Denmark, 8000 Aarhus C \\
	\texttt{kenneth.enevoldsen@cas.au.dk} \\
	\And
	\href{https://orcid.org/0000-0003-1113-4779}{\includegraphics[scale=0.06]{orcid.pdf}\hspace{1mm}Lasse~Hansen} \\
	Department of Clinical Medicine \&\\
	Center for Humanities Computing Aarhus\\
	Aarhus University\\
	Jens Chr. Skous Vej 4, Building 1483, 3rd floor\\
    Denmark, 8000 Aarhus C \\
	\texttt{lasse.hansen@clin.au.dk} \\
	\AND
	\href{https://orcid.org/0000-0002-5116-5070}{\includegraphics[scale=0.06]{orcid.pdf}\hspace{1mm}Kristoffer L.~Nielbo} \\
	Interacting Minds Centre \&\\
	Center for Humanities Computing Aarhus\\
	Aarhus University\\
	Jens Chr. Skous Vej 4, Building 1483, 3rd floor\\
    Denmark, 8000 Aarhus C \\
	\texttt{kln@cas.au.dk} \\
}

\date{}

\renewcommand{\headeright}{}
\renewcommand{\undertitle}{}
\renewcommand{\shorttitle}{DaCy: A Unified Framework for Danish NLP}

\hypersetup{
pdftitle={DaCy: A Unified Framework for Danish NLP},
pdfsubject={q-bio.NC, q-bio.QM},
pdfauthor={Kenneth C.~Enevoldsen, Lasse~Hansen, Kristoffer L.~Nielbo},
pdfkeywords={Natural Language Processing, Low-resource NLP, Data Augmentation, Danish NLP},
}

\begin{document}
\maketitle

\begin{abstract}
Danish natural language processing (NLP) has in recent years obtained considerable improvements with the addition of multiple new datasets and models. However, at present, there is no coherent framework for applying state-of-the-art models for Danish. We present DaCy: a unified framework for Danish NLP built on SpaCy. DaCy uses efficient multitask models which obtain state-of-the-art performance on named entity recognition, part-of-speech tagging, and dependency parsing. DaCy contains tools for easy integration of existing models such as for polarity, emotion, or subjectivity detection. In addition, we conduct a series of tests for biases and robustness of Danish NLP pipelines through augmentation of the test set of DaNE. DaCy large compares favorably and is especially robust to long input lengths and spelling variations and errors. All models except DaCy large display significant biases related to ethnicity while only Polyglot shows a significant gender bias. We argue that for languages with limited benchmark sets, data augmentation can be particularly useful for obtaining more realistic and fine-grained performance estimates. We provide a series of augmenters as a first step towards a more thorough evaluation of language models for low and medium resource languages and encourage further development.
\end{abstract}


\input{text/1_introduction}

\input{text/2_methods}

\input{text/3_experiments}

\input{text/4_results}

\input{text/5_discussion}

\printglossary[style=mcolindex, title=Abbreviations,nonumberlist]

\printbibliography






\end{document}

%% file: text/1_introduction.tex
\section{Introduction}

Danish \glsfirst{nlp} has seen a recent rise in resources with the introduction of the Danish Gigaword Corpus \parencite{derczynski_danish_2021}, curated lists of \gls{nlp} tools by DaNLP \parencite{brogaard_pauli_danlp_2021} and \url{sprogteknologi.dk}, and at least five pretrained neural language models \parencite{hojmark-bertelsen_aelaectra_2021, mollerhoj_danish_2019, tamini-sarnikowski_danish_2020}. Datasets and models are available for most common tasks such as \gls{ner}, \gls{pos} tagging, dependency parsing, sentiment analysis, and coreference resolution \parencite{brogaard_pauli_danlp_2021, sprogteknologi_2021}. However, no coherent, efficient and state-of-the-art framework exists for all fundamental \gls{nlp} tasks. Models are developed and distributed as disjoint projects and often require diverging package versions and have idiosyncratic APIs. These factors complicate workflows and hamper further developments.

\subsection{DaCy}
With this motivation we present DaCy: an efficient end-to-end framework for Danish \gls{nlp} with state-of-the-art performance on \gls{pos}, \gls{ner} and dependency parsing. DaCy fills the gap in Danish \gls{nlp} by providing a consistent interface that is easily extendable and able to integrate other models. DaCy is built on SpaCy v.3 which comes with a range of advantages: the framework is optimized, user-friendly, and well-documented. DaCy includes three fine-tuned language models: DaCy small, based on a Danish Electra (14M parameters) \parencite{hojmark-bertelsen_aelaectra_2021}; DaCy medium, based on the Danish BERT (110M parameters) \parencite{mollerhoj_danish_2019}; and DaCy large, based on the multilingual XLM-Roberta (550M parameters) \parencite{conneau_unsupervised_2020}. All models have been fine-tuned to do \gls{pos} tagging, \gls{ner}, and dependency parsing in a single forward pass, which increases the efficiency of the model and allows for larger models at the same computational cost.

Besides models fine-tuned for DaCy, the package includes convenient wrappers to add other models to the pipeline. For instance, Danish models for detecting polarity, emotion, and subjectivity classification can be added in a single line of code, and any HuggingFace Transformers \parencite{wolf_huggingfaces_2020} model trained for sentence classification can be conveniently wrapped and included in the pipeline using utility functions. With this functionality, DaCy aims at being a unified framework for Danish \gls{nlp}. All functionality is well-documented and covered by tutorials.\footnote{See: \url{https://centre-for-humanities-computing.github.io/DaCy/}}

\subsection{Robustness \& Evaluation}
Fine-tuned language models are commonly evaluated by testing performance on a gold-standard benchmark dataset. The most commonly used benchmark for Danish is the DaNE dataset \parencite{hvingelby_dane_2020}, which consists of the Danish Dependency Treebank \parencite{johannsen_universal_2015}, additionally tagged for \gls{ner}. For languages with few benchmarks datasets, such as Danish, the performance stability and generalizability can not be reliably estimated \parencite{ribeiro_beyond_2020}. For instance, the text included in DaNE was collected in the years 1983–1992 from both written and spoken domains \parencite{hvingelby_dane_2020}. Given the change of languages over time and the addition of new textual domains such as social media, this dataset is unlikely to be representative of the contemporary domains of application. For instance, models might not be sufficiently exposed to e.g. abbreviated names, spelling errors, or non-standard casing to correctly and robustly classify them. In this sense, the performance obtained on DaNE is unlikely to hold for real-world use cases.

To provide an additional layer of validation, we propose evaluating models on augmented gold-standard data. Data augmentation entails generating new data by slightly modifying existing data points \parencite{feng_survey_2021}. Data augmentation techniques such as rotation and cropping are widely used in computer vision to reduce overfitting \parencite{shorten_survey_2019}, and are becoming increasingly common in \gls{nlp} \parencite{chen_empirical_2021}. The complex syntactic and semantic structure of text complicates the task of finding useful augmentations, but simple manipulations such as synonym replacement and random character swaps and deletions have been found to be particularly useful for supervised learning in low-resource settings \parencite{wei_eda_2019}.

Although data augmentation is most commonly used for increasing the amount of training data, it can just as well be used for evaluation purposes \parencite{ribeiro_beyond_2020}. By augmenting a gold-standard dataset, we can evaluate model performance when exposed to data that more closely mimics real-life settings by adding spelling errors, more diverse names, or other manipulations. In \cref{sec:experiments}, we introduce a series of augmentations and evaluate the performance of Danish \gls{nlp} pipelines on them. 

The contributions of this paper are three-fold. 1) We introduce new state-of-art models for Danish dependency parsing, \gls{ner} and \gls{pos}. 2) We introduce the DaCy Python library as a unified framework for state-of-the-art \gls{nlp} in Danish. 3) We evaluate Danish \gls{nlp} pipelines using data augmentation and provide directions for future model development.

%% file: text/2_methods.tex
\section{Methods}

\subsection{Training}

To train the candidate models for DaCy, all publicly available language models for Danish were fine-tuned on the DaNE corpus \parencite{hvingelby_dane_2020} using SpaCy 3.0.3 \parencite{honnibal_spacy_2020}. The models include 2 Danish ELECTRAs \parencite{clark_electra_2020, hojmark-bertelsen_aelaectra_2021, tamini-sarnikowski_danish_2020}, the Danish ConvBERT \parencite{jiang_convbert_2021, tamini-sarnikowski_danish_2020}, the Danish BERT \parencite{devlin_bert_2019, mollerhoj_danish_2019}, and the multilingual XLM-Roberta Large \parencite{conneau_unsupervised_2020}. All models were trained with an input length of 10 sentences until convergence using similar hyperparameters on a Quadro RTX 8000 GPU. Adam was used as optimizer with hyperparameters $\beta_1= 0.9$ and $\beta_2 = 0.999$. Further, L2 normalization with $\alpha=0.01$ and gradient clipping with $c=1.0$ was employed. For increased efficiency, all models were trained with a multi-task objective \parencite{caruana_multitask_1997, ruder_overview_2017} on  \gls{ner}, \gls{pos}, and dependency parsing. This allows the training of larger models at the same computational cost, but it is unlikely that multi-task training at this scale improves performance \parencite{raffel_exploring_2020, aghajanyan_muppet_2021}.\footnote{For a full list of models and training configurations see the config files on Github: \url{https://github.com/centre-for-humanities-computing/DaCy/tree/main/training}}

\Cref{tab:dacy-perf} shows the performance of all fine-tuned models evaluated on DaNE's test set. The three best performing models in each size category, XLM-Roberta, DaBERT, and Ælæctra Cased are included in DaCy as the large, medium and small, respectively. In line with previous findings \parencite{raffel_exploring_2020, brown_language_2020, radford_language_2019}, larger models tend to perform better with XLM-Roberta obtaining the best performance across the board. 

\input{tables/dacy_performance}

%% file: tables/dacy_performance.tex
\begin{table}
\caption{\label{tab:dacy-perf}Performance of models finetuned for DaCy. Highest scores are in bold and second highest is underlined. WPS indicates words pr. second.}
\centering
\resizebox{\linewidth}{!}{
\begin{tabular}[t]{ll>{}c>{}c>{}c>{}c>{}c>{}c>{}c>{}c>{}c}
\toprule
\multicolumn{2}{c}{ } & \multicolumn{1}{c}{POS} & \multicolumn{5}{c}{NER} & \multicolumn{2}{c}{Dependency Parsing} & \multicolumn{1}{c}{Speed} \\
\cmidrule(l{3pt}r{3pt}){3-3} \cmidrule(l{3pt}r{3pt}){4-8} \cmidrule(l{3pt}r{3pt}){9-10} \cmidrule(l{3pt}r{3pt}){11-11}
Framework & Model & Accuracy & PER & LOC & ORG & MISC & Avg. F1 & UAS & LAS & WPS\\
\midrule
DaCy large & XLM-Roberta & \textbf{98.39} & \textbf{95.53} & \textbf{83.90} & \textbf{77.82} & \textbf{80.16} & \textbf{85.20} & \textbf{90.59} & \textbf{88} & 4311\\
DaCy medium & DaBERT & \underline{97.93} & \underline{89.62} & \underline{83.09} & \underline{67.35} & \underline{70.69} & \underline{78.47} & \underline{87.88} & \underline{85} & 8335\\
DaCy small & Ælæctra Cased & 97.69 & 87.36 & 81.95 & 63.83 & 70.68 & 76.55 & 86.45 & 83 & \textbf{10671}\\
\addlinespace
 & DaELECTRA & 97.40 & 82.80 & 77.39 & 63.01 & 66.95 & 73.16 & 85.20 & 82 & 9855\\
 & DaConvBERT & 97.23 & 85.08 & 78.26 & 61.76 & 66.93 & 73.77 & 84.61 & 81 & \underline{10029}\\
\bottomrule
\end{tabular}}
\end{table}

%% file: text/3_experiments.tex
\subsection{Evaluation}\label{sec:experiments}
To evaluate the robustness of DaCy and other Danish \gls{nlp} pipelines, we assessed their performance on multiple augmented version of the DaNE test set. All Danish models are trained on the DaNE corpus which consists of a mix of textual data of both spoken and written origin from the years 1983–1992 \parencite{hvingelby_dane_2020}, with the exception of Polyglot which is trained on entities extracted from Wikipedia \parencite{al-rfou_polyglot_2013}. As a consequence, the training data is rarely representative of the domain in which the models will be applied. For example, social media, contemporary news media, and historical texts have domain specific characteristics such as non-standard casing, a higher degree of typos, use of hashtags, and historic spelling such as upper-cased nouns \parencite{tahmasebi_study_2018, baldwin_social_2012, farzindar_natural_2015}. While it is infeasible to test the models on all possible domains, some of these characteristics can be modelled using data augmentation which can provide practitioners with an estimate of the potential shortcomings of the model. Further, data augmentation can be used to estimate biases against protected groups such as gender and ethnicity.

The augmenters presented here are not meant to be exhaustive, but rather a first step towards more thorough validation of new language models. We argue that the bar for inclusion of a new model should be set higher than a slight increase in benchmark performance. Language models are used in a variety of contexts which current benchmarks tasks, especially for low resource languages, do not capture. Our aim with these experiments is to provide an extra layer of insight into the performance of language models that more closely mimics naturalistic use cases, and encourage the development of further augmenters. Augmentation not only provides insights into when model performance breaks down, whether certain models are more suited for specific use-cases than others, but can also be used for identifying specific areas to improve upon.

The augmenters developed for this paper are designed in accordance with the SpaCy framework, and are thus not necessarily tied to DaCy or Danish in particular and can be used both during model validation and training. Comprehensive tutorials are provided on the DaCy Github repository.

We tested small, medium, and large SpaCy \parencite{honnibal_spacy_2020} and DaCy models, Stanza \parencite{qi_stanza_2020}, Polyglot \parencite{al-rfou_polyglot_2013}, NERDA \parencite{kjeldgaard_nerda_2020}, Flair \parencite{akbik_flair_2019}, and DaNLP's BERT \parencite{brogaard_pauli_danlp_2021} on the DaNE test set augmented with the following augmenters:

\begin{enumerate}
    \item Keystroke augmentation: substitute 2\%, 5\%, or 15\% of characters with a neighbouring character on a Danish QWERTY keyboard.
    \item ÆØÅ augmentation: substitute æ/Æ with ae/Ae, ø/Ø with oe/Oe, and å/Å with aa/Aa to simulate some historic text variations in Danish.
    \item Lower-case augmentation: convert all text to lower-case.
    \item Spacing augmentation: randomly remove 5\% of all whitespace.
    \item Name augmentations:
    \begin{enumerate}
        \item Substitute all names (PER entities) with randomly sampled Danish names, respecting first and last names.
        \item Substitute all names with randomly sampled names of Muslim origin used in Denmark \parencite{meldgaard_muslimske_2005}, respecting first and last names.
        \item Substitute all names with sampled Danish male names, respecting first and last names.
        \item Substitute all names with sampled Danish female names, respecting first and last names.
        \item Abbreviate all first names to the first character including a full stop.
    \end{enumerate}
\end{enumerate}

The stochastic augmentations, i.e. name and keystroke augmentations, were repeated 20 times.

Previous evaluations of Danish \gls{nlp} tools have used the gold-standard tokens instead of using a tokenization module. While this allows for easier comparison of the specific modules it inflates the performance metrics of the models and is unlikely to reflect the metric of interest, namely, the performance during application.\footnote{In our experiments, several of the Danish models performed worse using their own tokenizer.} All models were tested using both their own tokenizer (if they have one) and the SpaCy tokenizer for Danish. The performance reported in \cref{sec:results} uses the best peforming tokenization module for each pipeline. For all models except Stanza and Polyglot this was found to be the SpaCy tokenizer.

%% file: text/4_results.tex
\section{Results}\label{sec:results}

\input{tables/comparison}

\input{tables/ner}

\input{tables/pos}

\input{tables/dep}

\Cref{tab:all_together} shows the overall performance of Danish \gls{nlp} frameworks on \gls{pos}, \gls{ner}, and dependency parsing on the un-augmented DaNE test set. DaCy large obtains a new state-of-the-art on all tasks, most notably on \gls{ner} and dependency parsing. Regardless of model, performance for \gls{pos} is stable around 98\% accuracy. \gls{pos} tagging has long been at this level, and obtaining greater accuracy has been argued to require updates to the training data rather than new architectures \parencite{manning_part--speech_2011}.

\Cref{tab:aug_ner,tab:aug_pos,tab:aug_dep} shows a detailed performance breakdown of the models on \gls{ner}, \gls{pos}, and dependency parsing on the augmented data described in \cref{sec:experiments}. Overall, spelling variations and abbreviated first names consistently reduce performance of all models on all tasks. Even simple replacements of æ, ø, and å lead to performance degradation. In general, larger models handle augmentations better than small models with DaCy large performing the best on all augmentations with the exception of lower-casing. DaCy medium, DaNLP's BERT, and NERDA are based on the uncased Danish BERT \parencite{mollerhoj_danish_2019}, and are consequently not affected by casing. The BiLSTM-based models (Stanza and Flair) perform competitively under augmentations and are only consistently outperformed by DaCy large. 

On \gls{ner} specifically, all models with the exception of DaCy large obtain significantly worse performance on Muslim names as compared to Danish names. The robustness of DaCy large likely stems from the multilingual pre-training and the model size. Similarly, DaCy small is robust to spelling errors and outperforms larger models such as DaNLP's BERT and NERDA, this is likely due to its well-curated training data \parencite{derczynski_danish_2021}. DaNLP's BERT and NERDA models were found to severely under-perform if given longer input lengths. DaCy's models consistently perform slightly better with more context, but are not vulnerable to shorter input. Lastly, as expected, the lack of casing is especially detrimental for \gls{ner} for the cased models, most notably Flair, the SpaCy models, DaCy large and DaCy small.

%% file: tables/comparison.tex
\begin{table*}
\caption{\label{tab:all_together}Performance of Danish \gls{nlp} pipelines. Wall Time is the time taken by the model to go through the DaNE test set without augmentation. Stanza uses the spacy-stanza implementation. The speed of the DaNLP model is reported as provided by the framework, which does not utilize batch input. However, given the model size it can be expected to reach speeds comparable to DaCy medium. Empty cells indicates that the framework does not include the specific model.}
\centering
\resizebox{\linewidth}{!}{
\begin{tabular}[t]{l>{}c>{}c>{}c>{}c>{}c>{}c>{}c>{}c>{}cc}
\toprule
\multicolumn{1}{c}{ } & \multicolumn{1}{c}{POS} & \multicolumn{6}{c}{NER} & \multicolumn{2}{c}{Dependency Parsing} & \multicolumn{1}{c}{Wall Time} \\
\cmidrule(l{3pt}r{3pt}){2-2} \cmidrule(l{3pt}r{3pt}){3-8} \cmidrule(l{3pt}r{3pt}){9-10} \cmidrule(l{3pt}r{3pt}){11-11}
Model & Accuracy & Person & Location & Organization & Misc & F1 & F1 w/o Misc & LAS & UAS & GPU/CPU\\
\midrule
DaCy large & \textbf{98.37} & \textbf{93.33} & \textbf{84.88} & \textbf{76.49} & \textbf{80.16} & \textbf{84.39} & \textbf{85.65} & \textbf{88.44} & \textbf{90.85} & 2.9 / 34.7\\
DaCy medium & \underline{98.15} & 89.86 & 83.96 & 64.47 & 70.09 & 77.67 & 79.68 & \underline{86.65} & \underline{89.25} & 1.8 / 9.9\\
DaCy small & 97.75 & 87.98 & 79.23 & 60.58 & 64.82 & 74.18 & 76.98 & 84.03 & 87.63 & 1.9 / 2.6\\
\addlinespace
DaNLP BERT &  & 92.27 & 83.90 & \underline{71.13} &  & 72.84 & \underline{83.20} &  &  & 37.4 / -\\
Flair & 97.80 & \underline{92.60} & \underline{84.82} & 61.29 &  & 70.49 & 81.09 &  &  & 2.0 / -\\
NERDA &  & 92.35 & 81.52 & 65.96 & \underline{72.41} & \underline{79.04} & 80.85 &  &  & 2.5 / -\\
\addlinespace
Polyglot & 76.26 & 79.25 & 68.06 & 40.69 &  & 56.67 & 65.32 &  &  & - / 3.8\\
\addlinespace
SpaCy large & 96.30 & 86.17 & 84.16 & 63.36 & 65.52 & 75.75 & 78.57 & 78.01 & 81.95 & 0.9 / 1.4\\
SpaCy medium & 95.71 & 84.55 & 77.29 & 63.16 & 63.25 & 73.23 & 76.01 & 77.73 & 81.87 & 1.2 / 1.4\\
SpaCy small & 94.80 & 78.92 & 69.04 & 53.49 & 61.54 & 67.11 & 68.61 & 74.03 & 78.68 & 1.4 / 1.5\\
\addlinespace
Stanza & 97.62 &  &  &  &  &  &  & 83.84 & 87.34 & 29.3 / -\\
\bottomrule
\end{tabular}
}
\end{table*}

%% file: tables/ner.tex
\begin{table}
\caption{\label{tab:aug_ner}\gls{ner} performance of Danish \gls{nlp} pipelines reported as average F1 scores excluding the MISC category. Best scores are marked bold and second best are underlined. * denotes that the result is significantly different from the baseline using a significance threshold of 0.05 with Bonferroni correction for multiple comparisons. Danish names is considered the baseline for the augmentation of Muslim, female, and male names. Values in parentheses denote the standard deviation. NERDA limits input size to 128 wordpieces which leads to truncation on long input sizes and high rates of keystroke errors.}
\centering
\resizebox{\linewidth}{!}{
\begin{tabular}{>{\leavevmode\kern-\tabcolsep}p{7cm}cccccc<{\kern-\tabcolsep}}
\toprule
\multicolumn{2}{c}{ } & \multicolumn{5}{c}{Deterministic Augmentations} \\
\cmidrule(l{3pt}r{3pt}){3-7}
\multicolumn{4}{c}{ } & \multicolumn{2}{c}{Input Length} & \multicolumn{1}{c}{Names} \\
\cmidrule(lr{3pt}){5-6} \cmidrule(l{3pt}r{3pt}){7-7}
\hspace{0.3cm}Model & Baseline & Æøå & Lowercase & 5 sentences & 10 sentences & Abbreviated\\
\midrule
\hspace{0.3cm}DaCy large & \textbf{85.6} & \textbf{83.5} & 69.7 & \textbf{86.5} & \textbf{86.5} & \textbf{80.1}\\
\hspace{0.3cm}DaCy medium & 79.7 & 73.1 & 79.5 & 80.9 & 80.4 & 76.3\\
\hspace{0.3cm}DaCy small & 77.0 & 74.7 & 48.0 & 77.0 & 78.2 & 69.4\\
\addlinespace
\hspace{0.3cm}DaNLP BERT & \underline{83.2} & 78.6 & \textbf{83.1} & 78.6 & 61.9 & \underline{78.1}\\
\hspace{0.3cm}Flair & 81.1 & \underline{80.2} & 24.4 & \underline{81.0} & \underline{80.9} & 74.9\\
\hspace{0.3cm}NERDA & 80.9 & 74.8 & \underline{80.7} & 73.7 & 53.8 & 76.4\\
\addlinespace
\hspace{0.3cm}Polyglot & 65.3 & 61.4 & 55.3 & 64.8 & 64.2 & 40.2\\
\addlinespace
\hspace{0.3cm}SpaCy large & 78.6 & 75.4 & 5.7 & 78.8 & 78.8 & 78.0\\
\hspace{0.3cm}SpaCy medium & 76.0 & 74.7 & 9.7 & 76.5 & 76.8 & 76.0\\
\hspace{0.3cm}SpaCy small & 68.6 & 66.9 & 4.8 & 68.0 & 68.0 & 63.8\\
\bottomrule
\end{tabular}
}

\smallskip
\resizebox{\linewidth}{!}{
\tabcolsep=0.35cm
\begin{tabular}[t]{l>{}c>{}c>{}c>{}c>{}c>{}c>{}c}
\multicolumn{1}{c}{ } & \multicolumn{7}{c}{Stochastic Augmentations} \\
\cmidrule(l{3pt}r{3pt}){2-8}
\multicolumn{1}{c}{ } & \multicolumn{4}{c}{Names} & \multicolumn{3}{c}{Keystroke Errors} \\
\cmidrule(l{3pt}r{3pt}){2-5} \cmidrule(l{3pt}r{3pt}){6-8}
Model & Danish & Muslim & Female & Male & 2\% & 5\% & 15\%\\
\midrule
DaCy large & \textbf{86.2 (0.6)*} & \textbf{86.0 (0.5)} & \textbf{86.2 (0.5)} & \textbf{86.2 (0.4)} & \textbf{82.0 (1.2)*} & \textbf{76.9 (1.3)*} & \textbf{61.3 (1.6)*}\\
DaCy medium & 80.3 (0.5)* & 77.9 (0.8)* & 80.3 (0.4) & 80.2 (0.7) & 65.5 (1.7)* & 50.0 (1.6)* & 25.8 (1.3)*\\
DaCy small & 76.5 (0.9) & 75.7 (0.7)* & 76.7 (0.8) & 76.6 (0.7) & 70.7 (1.6)* & 62.1 (1.5)* & 41.3 (1.6)*\\
DaNLP BERT & \underline{82.9 (0.6)} & \underline{81.0 (1.0)*} & \underline{83.1 (0.5)} & \underline{83.0 (0.7)} & 72.6 (1.2)* & 60.9 (1.7)* & 37.0 (1.5)*\\
Flair & 81.2 (0.7) & 79.8 (0.7)* & 81.4 (0.5) & 81.5 (0.5) & \underline{78.3 (0.9)*} & \underline{73.5 (1.5)*} & \underline{56.3 (1.7)*}\\
\addlinespace
NERDA & 80.0 (1.1)* & 78.1 (1.2)* & 80.2 (0.8) & 80.0 (0.8) & 70.7 (1.4)* & 57.5 (1.4)* & 31.1 (1.6)*\\
Polyglot & 63.1 (1.2)* & 41.8 (0.7)* & 61.2 (1.2)* & 64.8 (1.2)* & 57.4 (0.9)* & 46.9 (1.9)* & 24.7 (1.9)*\\
SpaCy large & 79.5 (0.6)* & 71.6 (1.1)* & 79.8 (0.5) & 79.4 (0.5) & 72.1 (1.0)* & 63.3 (1.5)* & 44.9 (1.8)*\\
SpaCy medium & 78.2 (0.7)* & 69.2 (1.4)* & 78.2 (0.7) & 78.5 (0.8) & 70.5 (1.3)* & 64.2 (1.5)* & 46.9 (1.6)*\\
SpaCy small & 62.5 (1.6)* & 57.8 (1.4)* & 63.0 (1.1) & 63.3 (0.9) & 65.4 (0.7)* & 60.5 (1.5)* & 45.9 (1.6)*\\
\bottomrule
\end{tabular}
}

\end{table}

%% file: tables/pos.tex
\begin{table*}
\caption{\label{tab:aug_pos}POS performance of Danish \gls{nlp} pipelines reported as accuracy. Best scores are marked bold and second best are underlined. * denotes that the result is significantly different from baseline using a significance threshold of 0.05 with Bonferroni correction for multiple comparisons. Values in parentheses denote the standard deviation. NERDA limits input size to 128 wordpieces which leads to truncation on long input sizes and with a high degree of keystroke errors.}
\centering

\resizebox{\linewidth}{!}{
\begin{tabular}[t]{l>{}c>{}c>{}c>{}c>{}c>{}c>{}c>{}c}
\toprule
\multicolumn{2}{c}{ } & \multicolumn{4}{c}{Deterministic Augmentations} & \multicolumn{3}{c}{Stochastic Augmentations} \\
\cmidrule(l{3pt}r{3pt}){3-6} \cmidrule(l{3pt}r{3pt}){7-9}
\multicolumn{4}{c}{ } & \multicolumn{2}{c}{Input Length} & \multicolumn{3}{c}{Keystroke Errors} \\
\cmidrule(l{3pt}r{3pt}){5-6} \cmidrule(l{3pt}r{3pt}){7-9}
Model & Baseline & Æøå & Lowercase & 5 sentences & 10 sentences & 2\% & 5\% & 15\%\\
\midrule
DaCy large & \textbf{98.4} & \textbf{97.5} & \underline{95.5} & \textbf{98.5} & \textbf{98.4} & \textbf{95.5 (0.2)*} & \textbf{91.1 (0.2)*} & \underline{75.4 (0.6)*}\\
DaCy medium & \underline{98.2} & \underline{96.5} & \textbf{98.1} & \underline{97.8} & \underline{97.9} & 93.6 (0.3)* & 86.5 (0.3)* & 63.3 (0.6)*\\
DaCy small & 97.7 & 95.4 & 95.4 & 97.6 & 97.7 & 93.1 (0.2)* & 85.9 (0.4)* & 62.5 (0.4)*\\
\addlinespace
Flair & 97.8 & 95.0 & 95.0 & 97.7 & 97.7 & 94.7 (0.2)* & 89.8 (0.3)* & 72.1 (0.4)*\\
Polyglot & 76.3 & 71.6 & 75.6 & 75.7 & 75.6 & 71.7 (0.2)* & 65.3 (0.3)* & 49.4 (0.4)*\\
\addlinespace
SpaCy large & 96.3 & 92.4 & 91.5 & 96.3 & 96.3 & 91.5 (0.2)* & 84.8 (0.4)* & 66.2 (0.5)*\\
SpaCy medium & 95.7 & 92.4 & 91.6 & 95.8 & 95.7 & 91.0 (0.3)* & 84.5 (0.3)* & 66.0 (0.5)*\\
SpaCy small & 94.8 & 90.5 & 90.3 & 94.8 & 94.8 & 90.7 (0.2)* & 85.3 (0.3)* & 69.1 (0.4)*\\
\addlinespace
Stanza & 97.6 & 96.1 & 95.4 & 97.7 & 97.7 & \underline{94.8 (0.2)*} & \underline{90.6 (0.3)*} & \textbf{75.6 (0.5)*}\\
\bottomrule
\end{tabular}
}
\end{table*}

%% file: tables/dep.tex
\begin{table}
\caption{\label{tab:aug_dep}Dependency parsing performance of Danish \gls{nlp} pipelines reported as LAS. Best scores are marked bold and second best are underlined. * denotes that the result is significantly different from baseline using a significance threshold of 0.05 with Bonferroni correction for multiple comparisons. Values in parentheses denote the standard deviation.}
\centering
\resizebox{\linewidth}{!}{
\begin{tabular}[t]{l>{}c>{}c>{}c>{}c>{}c>{}c>{}l>{}c}
\toprule
\multicolumn{2}{c}{ } & \multicolumn{4}{c}{Deterministic Augmentations} & \multicolumn{3}{c}{Stochastic Augmentations} \\
\cmidrule(l{3pt}r{3pt}){3-6} \cmidrule(l{3pt}r{3pt}){7-9}
\multicolumn{4}{c}{ } & \multicolumn{2}{c}{Input Length} & \multicolumn{3}{c}{Keystroke Errors} \\
\cmidrule(l{3pt}r{3pt}){5-6} \cmidrule(l{3pt}r{3pt}){7-9}
Model & Baseline & Æøå & Lowercase & 5 sentences & 10 sentences & 2\% & 5\% & 15\%\\
\midrule
DaCy large & \textbf{88.4} & \textbf{86.2} & \textbf{87.0} & \textbf{88.3} & \textbf{88.3} & \textbf{83.7 (0.4)*} & \textbf{76.6 (0.5)*} & \textbf{53.6 (0.8)*}\\
DaCy medium & \underline{86.7} & \underline{84.6} & \underline{86.6} & \underline{85.4} & \underline{85.3} & \underline{79.9 (0.5)*} & 69.9 (0.7)* & 41.1 (0.9)*\\
DaCy small & 84.0 & 79.0 & 82.7 & 83.5 & 83.0 & 76.8 (0.4)* & 66.2 (0.8)* & 38.0 (0.6)*\\
\addlinespace
SpaCy large & 78.0 & 71.0 & 74.0 & 77.6 & 77.6 & 69.7 (0.5)* & 59.3 (0.7)* & 34.8 (0.7)*\\
SpaCy medium & 77.7 & 71.2 & 73.8 & 77.4 & 77.4 & 69.6 (0.6)* & 59.5 (0.6)* & 35.3 (0.7)*\\
SpaCy small & 74.0 & 65.9 & 70.4 & 74.1 & 74.1 & 67.5 (0.4)* & 59.1 (0.5)* & 38.2 (0.7)*\\
\addlinespace
Stanza & 83.8 & 80.2 & 82.5 & 83.9 & 83.9 & 79.0 (0.4)* & \underline{71.9 (0.5)*} & \underline{49.8 (0.9)*}\\
\bottomrule
\end{tabular}}
\end{table}

%% file: text/5_discussion.tex
\section{Discussion}
This paper has introduced the DaCy models and presented a thorough evaluation of Danish \gls{nlp} models on a battery of augmentations. DaCy models achieve state-of-the-art performance on Danish \gls{ner}, \gls{pos}, and dependency parsing, and are robust to augmentations such as keystroke errors, name changes, and lowercasing. The results from training DaCy underline three well-known trends in deep learning and \gls{nlp},
1) larger models tend to perform better, 2) higher quality pre-training data leads to better models, as illustrated by the superior performance of Ælæctra compared to DaELECTRA, and 3) multilingual models perform competitively with monolingual models \parencite{raffel_exploring_2020, xue_mt5_2021, brown_language_2020}.

Our experiments with multiple augmenters revealed different patterns of strengths and weaknesses across Danish NLP models. In general, larger models tend to be more robust to data augmentations. Several models are highly sensitive to casing, which limits their usefulness on certain domains. Evaluating models on augmented data provides a more holistic and realistic estimate of the expected performance, and can reveal in which use cases one model might be more useful than another. For example, it might be better to use DaCy medium on social media as opposed to DaCy large as its performance is not affected by casing. 

The purpose of the data augmentation experiments was to evaluate the robustness of Danish models and to open a discussion on how to present new models going forward. As more models are developed for low and medium resource languages, properly evaluating them becomes vital for securing robustness, transparency, and effectiveness despite limited benchmark sets. We do not posit data augmentation as the only solution, but demonstrate that it can effectively reveal performance differences on important factors such as casing, spelling errors, and biases related to protected groups. As researchers, we bear the responsibility for releasing adequately tested and robust models into the world. With the increasing ease of deployment, users must be made aware of the level of performance they can realistically expect to achieve on their problem, and when to choose one model over another. Social media researchers should know that certain models are sensitive to casing, historians should know that some models handle old text variations such as ae, oe, aa poorly, and lawyers should be aware that models might not be able to identify abbreviated names as effectively. In this regard, transparency and openness as to when and how models fail are crucial measures to report. Such evaluation requires the development of infrastructure and tools, but is fast and easy to conduct once in place. For instance, it only takes 8 minutes to test DaCy large on all augmented datasets including bootstrapping. As part of the DaCy library, we provide several augmenters and utility functions for evaluation that integrate with SpaCy and encourage new \gls{nlp} models to use and expand upon them. For the continued development of low and medium resource \gls{nlp} in a direction that is beneficial for practitioners, it is vital to conduct more thorough evaluation of new models. We suggest these augmenters not as an evaluation standard, but as preliminary guiding principles for future development of \gls{nlp} models for low and medium resource languages in particular.